\documentclass[11pt,a4paper]{article}

\usepackage{acl2015}

\usepackage{times}
\usepackage{latexsym}
\usepackage{amsmath}
\usepackage{multirow}
\usepackage{url}

\usepackage{mylingmacros}
\usepackage{amsmath}
\usepackage{multirow}
\usepackage{graphicx}
\usepackage{tikz}
\usepackage{tikz-qtree}
\usepackage{algorithm}
\usepackage{algorithmic}

\newcommand{\nlang}[1]{{\it #1}}

\title{Unsupervised Sentence Simplification Using Deep Semantics}

\author{Shashi Narayan\\
School of Informatics\\
The University of Edinburgh\\
Edinburgh, EH8 9AB, UK\\
{\tt shashi.narayan@ed.ac.uk} \\\And
Claire Gardent\\
CNRS, LORIA, UMR 7503\\
Vandoeuvre-l\`es-Nancy, F-54500, France\\
{\tt claire.gardent@loria.fr}}

\begin{document}
\maketitle
\begin{abstract}

We present a novel approach to sentence simplification which departs
from previous work in two main ways. First, it requires neither hand
written rules nor a training corpus of aligned standard and simplified
sentences. Second, sentence splitting operates on deep semantic
structure. We show (i) that the unsupervised framework we propose is
competitive with four state-of-the-art supervised systems and (ii)
that our semantic based approach allows for a principled and effective
handling of sentence splitting.


\end{abstract}

\section{Introduction}
\label{sec:introduction}

Sentence simplification maps a sentence to a simpler, more readable
one approximating its content.
As has been argued in \cite{shardlow2014survey},
sentence simplification has many potential applications. It is useful
as a preprocessing step for a variety of NLP systems such as parsers
and machine translation systems \cite{chandrasekar1996motivations},
summarisation \cite{knight2000statistics}, sentence fusion
\cite{filippova2008dependency} and semantic role labelling
\cite{vickrey2008sentence}. It also has wide ranging potential
societal applications as a reading aid for people with aphasia
\cite{carroll1999simplifying}, for low literacy readers
\cite{watanabe2009facilita} and for non native speakers
\cite{siddharthan2002architecture}.


In this paper, we present a novel approach to sentence simplification
which departs from previous work in two main ways.  First, it requires
neither hand written rules nor a training corpus of aligned standard
and simplified sentences. Instead, we exploit non aligned Simple and
English Wikipedia to learn the probability of lexical simplifications,
of the semantics of simple sentences and of optional phrases i.e.,
phrase which may be deleted when simplifying.  Second, sentence
splitting is semantic based.  We show (i) that our unsupervised
framework is competitive with four state-of-the-art systems and (ii)
that our semantic based approach allows for a principled and effective
handling of sentence splitting.

\section{Related Work}
\label{sec:related}

Earlier work on sentence simplification relied on handcrafted rules to
capture syntactic simplification e.g., to split coordinated and
subordinated sentences into several, simpler clauses or to model e.g.,
active/passive transformations
\cite{siddharthan2002architecture,chandrasekar1997automatic,canning2002syntactic,siddharthan2011text,siddharthan2010complex}.
While these hand-crafted approaches can encode precise and
linguistically well-informed syntactic transformations, they do not
account for lexical simplifications and their interaction with the
sentential context. \newcite{siddharthan-mandya:2014:EACL} therefore
propose an approach where hand-crafted syntactic simplification rules
are combined with lexical simplification rules extracted from aligned
English and simple English sentences, and revision histories of Simple
Wikipedia.

Using the parallel dataset formed by Simple English Wikipedia
(SWKP)\footnote{\url{http://simple.wikipedia.org}} 
and traditional English Wikipedia
(EWKP)\footnote{\url{http://en.wikipedia.org}}, further work has
focused on developing machine learning approaches to sentence
simplification.

\newcite{zhu2010monolingual} constructed a parallel Wikipedia corpus
(PWKP) of 108,016/114,924 complex/simple sentences by aligning
sentences from EWKP and SWKP and used the resulting bitext to train a
simplification model inspired by \emph{syntax-based} machine
translation \cite{yamada2001syntax}. Their simplification model
encodes the probabilities for four rewriting operations on the parse
tree of an input sentences namely, substitution, reordering, splitting
and deletion. It is combined with a language model to improve
grammaticality and the decoder translates sentences into simpler ones
by greedily selecting the output sentence with highest probability.

Using both the PWKP corpus developed by \newcite{zhu2010monolingual}
and the edit history of simple Wikipedia,
\newcite{woodsend2011learning} learn a quasi synchronous grammar
\cite{smith2006quasi} describing a loose alignment between parse trees
of complex and of simple sentences. Following \newcite{dras1999tree},
they then generate all possible rewrites for a source tree and use
integer linear programming to select the most appropriate
simplification. They evaluate their model on the same dataset used by
\newcite{zhu2010monolingual} namely, an aligned corpus of 100/131
EWKP/SWKP sentences.

\newcite{wubben2012sentence}, \newcite{coster2011learning} and
\newcite{wei16tacl} saw simplification as a monolingual translation
task where the complex sentence is the source and the simpler one is
the target. To account for deletions, reordering and substitution,
\newcite{coster2011learning} trained a phrase based machine
translation system on the PWKP corpus while modifying the word
alignment output by GIZA++ in Moses to allow for null phrasal
alignments. In this way, they allow for phrases to be deleted during
translation. Similarly, \newcite{wubben2012sentence} used Moses and
the PWKP data to train a phrase based machine translation system
augmented with a post-hoc reranking procedure designed to rank the
output based on their dissimilarity from the source sentence. Unlinke
\newcite{wubben2012sentence} and \newcite{coster2011learning} who used
machine translation as a black box, \newcite{wei16tacl} proposed to
modify the optimization function of SMT systems by tuning them for the
sentence simplification task. However, in their work they primarily
focus on lexical simplification.

Finally, \newcite{narayan2014hybrid} present a hybrid approach
combining a probabilistic model for sentence splitting and deletion
with a statistical machine translation system trained on PWKP for
substitution and reordering. 

Our proposal differs from all these approaches in that it does not use
the parallel PWKP corpus for training. Nor do we use hand-written
rules. Another difference is that we use a deep semantic
representation as input for simplification. While a similar approach
was proposed in \cite{narayan2014hybrid}, the probabilistic models
differ in that we determine splitting points based on the maximum
likelihood of sequences of thematic role sets present in SWKP whereas
\newcite{narayan2014hybrid} derive the probability of a split from the
aligned EWKP/SWKP corpus using expectation maximisation. As we shall
see in Section~\ref{sec:xps}, because their data is more sparse,
\newcite{narayan2014hybrid} predicts less and lower quality
simplifications by sentence splitting.

\section{Simplification Framework}
\label{sec:fwk}

Our simplification framework pipelines three dedicated modules
inspired from previous work on lexical simplification, syntactic
simplification and sentence compression. All three modules are
unsupervised.

\begin{figure*}[htbp]
  \center{
    \footnotesize
    \begin{tikzpicture}[scale=1.0]
      \begin{scope}[shift={(0,2)}]
        \draw (7.5,0) node {$\text{In 1964 Peter Higgs published his
            second paper in Physical Review Letters describing
            Higgs mechanism which predicted}$};
        \draw (7.5,-0.4) node {$\text{a new massive spin-zero boson
            for the first time .}$};
        \draw (8, -1) node {$\Bigg\Downarrow \textbf{Lex Simpl.}$};
        \draw (7.5,-1.7) node {$\text{In 1964 Peter Higgs wrote his
            second paper in Physical Review Letters explaining Higgs
            mechanism which predicted}$}; 
        \draw (7.5,-2.1) node {$\text{a new massive elementary
            particle for the first time .}$};
      \end{scope}

      \draw (0,-1.5) node {$($};
      \draw (0.15,-1.5) node {$($};
      \begin{scope}[shift={(0.3,-2.25)}]
        \draw (0,0) -- (0, 1.5) -- (3,1.5) -- (3,0) -- (0,0);
        \draw (0,1) -- (3,1);
        \draw (0.5,1.25) node {$X_0$};
        \draw (1.5,0.75) node {$\text{named}(X_0, \text{higgs}, per)$};
        \draw (1.5,0.25) node {$\text{named}(X_0, \text{peter}, per)$};
      \end{scope}
      \draw (3.5,-1.5) node {$\wedge$};
      \draw (3.65,-1.5) node {$($};
      \begin{scope}[shift={(3.8,-2)}]
        \draw (0,0) -- (0, 1) -- (2,1) -- (2,0) -- (0,0);
        \draw (0,0.5) -- (2,0.5);
        \draw (0.5,0.75) node {$X_1$};
        \draw (1,0.25) node {$\text{male}(X_1)$};
      \end{scope}
      \draw (6,-1.5) node {$\wedge$};
      \draw (6.15,-1.5) node {$($};
      \begin{scope}[shift={(6.3,-2.5)}]
        \draw (0,0) -- (0, 2) -- (2,2) -- (2,0) -- (0,0);
        \draw (0,1.5) -- (2,1.5);
        \draw (0.5,1.75) node {$X_2$};
        \draw (1,1.25) node {$\text{second}(X_2)$};
        \draw (1,0.75) node {$\text{paper}(X_2)$};
        \draw (1,0.25) node {$\text{of}(X_2, X_1)$};
      \end{scope}
      \draw (8.5,-1.5) node {$\wedge$};
      \draw (8.65,-1.5) node {$($};
      \begin{scope}[shift={(8.8,-2.5)}]
        \draw (0,0) -- (0, 2) -- (2.5,2) -- (2.5,0) -- (0,0);
        \draw (0,1.5) -- (2.5,1.5);
        \draw (0.5,1.75) node {$X_3$};
        \draw (1.25,1.25) node {$\text{write}(X_3)$};
        \draw (1.25,0.75) node {$\text{agent}(X_3, X_0)$};
        \draw (1.25,0.25) node {$\text{patient}(X_3, X_2)$};
      \end{scope}  
      \draw (11.45,-1.5) node {$;$};
      \draw (11.6,-1.5) node {$($};
      \begin{scope}[shift={(11.75,-2.5)}]
        \draw (0,0) -- (0, 2) -- (3.5,2) -- (3.5,0) -- (0,0);
        \draw (0,1.5) -- (3.5,1.5);
        \draw (0.5,1.75) node {$X_4$};
        \draw (1.75,1.25) node {$\text{named}(X_4, \text{physical}, org)$};
        \draw (1.75,0.75) node {$\text{named}(X_4, \text{review}, org)$};
        \draw (1.75,0.25) node {$\text{named}(X_4, \text{letters}, org)$};
      \end{scope}
      \draw (15.45,-1.5) node {$\wedge$};
      \begin{scope}[shift={(0,-5.8)}]
        \draw (0,0) -- (0, 3) -- (2.5,3) -- (2.5,0) -- (0,0);
        \draw (0,2.5) -- (2.5,2.5);
        \draw (0.5,2.75) node {$X_5$};
        \draw (1.25,2.25) node {$\text{thing}(X_5)$};
        \draw (1.25,1.75) node {$\text{event}(X_3)$};
        \draw (1.25,1.25) node {$\text{in}(X_3, X_4)$};
        \draw (1.25,0.75) node {$\text{in}(X_3, X_5)$};
        \draw (1.25,0.25) node {$\text{timex}(X_5)=\text{1964}$};
      \end{scope}  
      \draw (2.65, -4.25) node {$)$};
      \draw (2.8, -4.25) node {$)$};
      \draw (2.95, -4.25) node {$)$};
      \draw (3.1, -4.25) node {$)$};
      \draw (3.25, -4.25) node {$)$};
      \draw (3.4, -4.25) node {$;$};
      \draw (3.55, -4.25) node {$($};
      \begin{scope}[shift={(3.7,-4.75)}]
        \draw (0,0) -- (0, 1) -- (2,1) -- (2,0) -- (0,0);
        \draw (0,0.5) -- (2,0.5);
        \draw (0.5,0.75) node {$X_6$};
      \end{scope}
      \draw (5.85, -4.25) node {$;$};
      \draw (6, -4.25) node {$($};
      \begin{scope}[shift={(6.15,-5.25)}]
        \draw (0,0) -- (0, 2) -- (3,2) -- (3,0) -- (0,0);
        \draw (0,1.5) -- (3,1.5);
        \draw (0.75,1.75) node {$X_7, X_8$};
        \draw (1.5,1.25) node {$\text{mechanism}(X_8)$};
        \draw (1.5,0.75) node {$\text{nn}(X_7, X_8)$};
        \draw (1.5,0.25) node {$\text{named}(X_7, \text{higgs}, org)$};
      \end{scope}
      \draw (9.35,-4.25) node {$\wedge$};
      \begin{scope}[shift={(9.55,-5.75)}]
        \draw (0,0) -- (0, 3) -- (6,3) -- (6,0) -- (0,0);
        \draw (0,2.5) -- (6,2.5);
        \draw (1.5,2.75) node {$X_9, X_{10}, X_{11}, X_{12}$};
        \draw (1,2.25) node {$\text{new}(X_9)$};
        \draw (1,1.75) node {$\text{massive}(X_9)$};
        \draw (1,1.25) node {$\text{elementary}(X_9)$};
        \draw (1,0.75) node {$\text{particle}(X_9)$};
        \draw (1,0.25) node {$\text{predict}(X_{10})$};
        \draw (2.8,2.25) node {$\text{event}(X_{10})$};
        \draw (2.8,1.75) node {$\text{explain}(X_{11})$};
        \draw (2.8,1.25) node {$\text{event}(X_{11})$};
        \draw (2.8,0.75) node {$\text{first}(X_{12})$};
        \draw (2.8,0.25) node {$\text{time}(X_{12})$};
        \draw (4.85,2.25) node {$\text{agent}(X_{10}, X_8)$};
        \draw (4.85,1.75) node {$\text{patient}(X_{10}, X_9)$};
        \draw (4.85,1.25) node {$\text{agent}(X_{11}, X_6)$};
        \draw (4.85,0.75) node {$\text{patient}(X_{11}, X_8)$};
        \draw (4.85,0.25) node {$\text{for}(X_{10}, X_{12})$};
      \end{scope}
      \draw (15.65, -4.25) node {$)$};
      \draw (15.8, -4.25) node {$)$};
      \draw (6, -6.15) node {[Discourse Representation Structure produced by BOXER]};

      \begin{scope}[shift={(6,-6.75)}, level distance=1cm, sibling distance=0.5cm] 
        \Tree
            [.$\text{ROOT}$
              \edge node[auto=right]{$$};
                    [.$X_3$
                      \edge node[auto=right]{$R_1$};
                      $X_0$
                      \edge node[auto=right]{$R_2$};
                            [.$X_2$
                              \edge node[auto=right]{$R_3$};
                              $X_1$
                            ]
                      \edge node[auto=right]{$R_4$};
                      $X_4$
                      \edge node[auto=left]{$R_5$};
                      $X_5$
                    ]
              \edge node[auto=right]{$$};
                    [.$X_{11}$
                      \edge node[auto=right]{$R_6$};
                      $X_6$
                      \edge node[auto=left]{$R_7$};
                            [.\node (X8) {$X_8$};
                              \edge node[auto=right]{$R_8$};
                              $X_7$
                            ]
                    ] 
              \edge node[auto=right]{$$};
                    [.\node (X10) {$X_{10}$};
                      \edge node[auto=left]{$R_{10}$};
                      $X_9$
                      \edge node[auto=left]{$R_{11}$};
                      $X_{12}$
                    ]                    
              \edge node[auto=right]{$$};
              $O_1$
            ]
            \draw (X8.north) -- node[left] {$R_9$} (X10.south);
            \draw (-2.1, -2.9) node {[DRS Graph Representation]};
      \end{scope} 

      \begin{scope}[shift={(10.5,-14.5)}] 
        \draw (0,0) -- (0, 8.5) -- (5,8.5) -- (5,0) -- (0,0);
        \draw[dotted] (0.75,0) -- (0.75, 8.5);
        \draw[dotted] (2,0) -- (2, 8.5);
        
        \draw (0.375,0.25) node {$O_1$};
        \draw (1.375,0.25) node {$16$};
        \draw (3.5,0.25) node {$\text{which/WDT}$};
        \draw[dotted] (0,0.5) -- (5, 0.5);

        \draw (0.375,0.75) node {$X_{12}$};
        \draw (1.375,0.75) node {$24,25,26$};
        \draw (3.5,0.75) node {$\text{first/a}, \text{time/n}$};
        \draw[dotted] (0,1) -- (5, 1);

        \draw (0.375,1.25) node {$X_{11}$};
        \draw (1.375,1.25) node {$13$};
        \draw (3.5,1.25) node {$\text{explain/v}, \textbf{event}$};
        \draw[dotted] (0,1.5) -- (5, 1.5);
        
        \draw (0.375,1.75) node {$X_{10}$};
        \draw (1.375,1.75) node {$17$};
        \draw (3.5,1.75) node {$\text{predict/v}, \textbf{event}$};
        \draw[dotted] (0,2) -- (5, 2);

        \draw (0.375,2.5) node {$X_{9}$};
        \draw (1.375,2.75) node {$18,19,20$};
        \draw (1.375,2.25) node {$21,22$};
        \draw (3.5,2.75) node {$\text{new/a}, \text{elementary/a}$};
        \draw (3.5,2.25) node {$\text{massive/a}, \text{particle/n}$};
        \draw[dotted] (0,3) -- (5, 3);

        \draw (0.375,3.25) node {$X_8$};
        \draw (1.375,3.25) node {$14,15$};
        \draw (3.5,3.25) node {$\text{mechanism/n}$};
        \draw[dotted] (0,3.5) -- (5, 3.5);

        \draw (0.375,3.75) node {$X_7$};
        \draw (1.375,3.75) node {$14$};
        \draw (3.5,3.75) node {$\text{higgs/org}$};
        \draw[dotted] (0,4) -- (5, 4);
        
        \draw (0.375,4.25) node {$X_6$};
        \draw (1.375,4.25) node {$6,7,8$};
        \draw (3.5,4.25) node {$--$};
        \draw[dotted] (0,4.5) -- (5, 4.5);
        
        \draw (0.375,4.75) node {$X_5$};
        \draw (1.375,4.75) node {$2$};
        \draw (3.5,4.75) node {$\text{thing/n}, \text{1964}$};
        \draw[dotted] (0,5) -- (5, 5);
        
        \draw (0.375,5.5) node {$X_4$};
        \draw (1.375,5.5) node {$10,11,12$};
        \draw (3.5,5.75) node {$\text{physical/org}$};
        \draw (3.5,5.25) node {$\text{review/org},\text{letters/org}$};
        \draw[dotted] (0,6) -- (5, 6);

        \draw (0.375,6.25) node {$X_3$};
        \draw (1.375,6.25) node {$5$};
        \draw (3.5,6.25) node {$\text{write/v}, \textbf{event}$};
        \draw[dotted] (0,6.5) -- (5, 6.5);

        \draw (0.375,6.75) node {$X_2$};
        \draw (1.375,6.75) node {$6,7,8$};
        \draw (3.5,6.75) node {$\text{second/a}, \text{paper/n}$};
        \draw[dotted] (0,7) -- (5, 7);

        \draw (0.375,7.25) node {$X_1$};
        \draw (1.375,7.25) node {$6$};
        \draw (3.5,7.25) node {$\text{male/a}$};
        \draw[dotted] (0,7.5) -- (5, 7.5);

        \draw (0.375,7.75) node {$X_0$};
        \draw (1.375,7.75) node {$3,4$};
        \draw (3.5,7.75) node {$\text{higgs/per}, \text{peter/per}$};
        \draw[dotted] (0,8) -- (5,8);
        
        \draw (0.375,8.25) node {$\text{node}$};
        \draw (1.375,8.25) node {$\text{pos. in S}$};
        \draw (3.5,8.25) node {$\text{predicate/type}$};
        \draw[dotted] (0,8.5) -- (5,8.5);
      \end{scope} 
        
      \begin{scope}[shift={(10.5,-20.7)}] 
        \draw (0,0) -- (0, 6) -- (5,6) -- (5,0) -- (0,0);
        \draw[dotted] (0.75,0) -- (0.75, 6);
        \draw[dotted] (2,0) -- (2, 6);
        
        \draw (0.375,0.25) node {$R_{11}$};
        \draw (1.375,0.25) node {$23$};
        \draw (3.5,0.25) node {$for, X_{10} \rightarrow X_{12}$};
        \draw[dotted] (0,0.5) -- (5, 0.5);

        \draw (0.375,0.75) node {$R_{10}$};
        \draw (1.375,0.75) node {$17$};
        \draw (3.5,0.75) node {$patient, X_{10} \rightarrow X_9$};
        \draw[dotted] (0,0.5) -- (5, 0.5);

        \draw (0.375,1.25) node {$R_9$};
        \draw (1.375,1.25) node {$17$};
        \draw (3.5,1.25) node {$agent, X_{10} \rightarrow X_8$};
        \draw[dotted] (0,1) -- (5, 1);
        
        \draw (0.375,1.75) node {$R_8$};
        \draw (1.375,1.75) node {$--$};
        \draw (3.5,1.75) node {$nn, X_8 \rightarrow X_7$};
        \draw[dotted] (0,1.5) -- (5, 1.5);
        
        \draw (0.375,2.25) node {$R_7$};
        \draw (1.375,2.25) node {$13$};
        \draw (3.5,2.25) node {$patient, X_{11} \rightarrow X_8$};
        \draw[dotted] (0,2) -- (5, 2);

        \draw (0.375,2.75) node {$R_6$};
        \draw (1.375,2.75) node {$13$};
        \draw (3.5,2.75) node {$agent, X_{11} \rightarrow X_6$};
        \draw[dotted] (0,2.5) -- (5, 2.5);

        \draw (0.375,3.25) node {$R_5$};
        \draw (1.375,3.25) node {$1$};
        \draw (3.5,3.25) node {$in, X_3 \rightarrow X_5$};
        \draw[dotted] (0,3) -- (5, 3);

        \draw (0.375,3.75) node {$R_4$};
        \draw (1.375,3.75) node {$9$};
        \draw (3.5,3.75) node {$in, X_3 \rightarrow X_4$};
        \draw[dotted] (0,3.5) -- (5, 3.5);
        
        \draw (0.375,4.25) node {$R_3$};
        \draw (1.375,4.25) node {$6$};
        \draw (3.5,4.25) node {$of, X_2 \rightarrow X_1$};
        \draw[dotted] (0,4) -- (5, 4);

        \draw (0.375,4.75) node {$R_{2}$};
        \draw (1.375,4.75) node {$5$};
        \draw (3.5,4.75) node {$patient, X_3 \rightarrow X_2$};
        \draw[dotted] (0,4.5) -- (5, 4.5);

        \draw (0.375,5.25) node {$R_1$};
        \draw (1.375,5.25) node {$5$};
        \draw (3.5,5.25) node {$agent, X_3 \rightarrow X_0$};
        \draw[dotted] (0,5) -- (5, 5);

        \draw (0.375,5.75) node {rel};
        \draw (1.375,5.75) node {pos. in S};
        \draw (3.5,5.75) node {predicate};
        \draw[dotted] (0,5.5) -- (5, 5.5);
      \end{scope} 
      
      \begin{scope}[shift={(3.3,-11)}, level distance=1cm, sibling distance=0.4cm] 
        \Tree
            [.$\text{ROOT}$
              \edge node[auto=right]{$$};
                    [.$X_3$
                      \edge node[auto=right]{$R_1$};
                      $X_0$
                      \edge node[auto=right]{$R_2$};
                            [.$X_2$
                              \edge node[auto=right]{$R_3$};
                              $X_1$
                            ]
                      \edge node[auto=right]{$R_4$};
                      $X_4$
                      \edge node[auto=left]{$R_5$};
                      $X_5$
                    ]
              \edge node[auto=right]{$$};
                    [.$X_{11}$
                      \edge node[auto=right]{$R_6$};
                      $X_6$
                      \edge node[auto=left]{$R_7$};
                            [.$X_8$
                              \edge node[auto=right]{$R_8$};
                              $X_7$
                            ]
                    ]                
            ]
      \end{scope}
      \begin{scope}[shift={(8.5,-11)}, level distance=1cm, sibling distance=0.4cm] 
        \Tree
            [.$\text{ROOT}$
              \edge node[auto=right]{$$};
                    [.$X_{10}$
                      \edge node[auto=right]{$R_9$};
                            [.$X_8$
                              \edge node[auto=left]{$R_8$};
                              $X_7$
                            ]
                      \edge node[auto=right]{$R_{10}$};
                      $X_9$
                      \edge node[auto=left]{$R_{11}$};
                      $X_{12}$
                    ]     
            ]
      \end{scope}
      \begin{scope}[shift={(0,-12)}]
        \draw (0, 0) node {$($};
        \draw (10.2, 0) node {$)$};
        \draw (6.5, 1.5) node {$\Bigg\Downarrow \textbf{Split}$};
      \end{scope}

      \begin{scope}[shift={(3.3,-15.5)}, level distance=1cm, sibling distance=0.4cm] 
        \Tree
            [.$\text{ROOT}$
              \edge node[auto=right]{$$};
                    [.$X_3$
                      \edge node[auto=right]{$R_1$};
                      $X_0$
                      \edge node[auto=right]{$R_2$};
                            [.$X_2^{'}$
                              \edge node[auto=right]{$R_3$};
                              $X_1$
                            ]
                      \edge node[auto=left]{$R_5$};
                      $X_5$
                    ]
              \edge node[auto=right]{$$};
                    [.$X_{11}$
                      \edge node[auto=right]{$R_6$};
                      $X_6$
                      \edge node[auto=left]{$R_7$};
                            [.$X_8$
                              \edge node[auto=right]{$R_8$};
                              $X_7$
                            ]
                    ]                
            ]

            \draw (0,-3.5) node {$\text{In 1964 Peter Higgs wrote his}$};
            \draw (0,-3.9) node {$\text{paper explaining Higgs mechanism.}$};
      \end{scope}
      \begin{scope}[shift={(8.5,-15.5)}, level distance=1cm, sibling distance=0.4cm] 
        \Tree
            [.$\text{ROOT}$
              \edge node[auto=right]{$$};
                    [.$X_{10}$
                      \edge node[auto=right]{$R_9$};
                            [.$X_8$
                              \edge node[auto=left]{$R_8$};
                              $X_7$
                            ]
                      \edge node[auto=left]{$R_{10}$};
                      $X_9^{'}$
                    ]     
            ]
            \draw (0,-3.5) node {$\text{Higgs mechanism predicted}$};
            \draw (0,-3.9) node {$\text{a new elementary particle.}$};
      \end{scope}
      \begin{scope}[shift={(0,-16.8)}]
        \draw (0, 0) node {$($};
        \draw (10.2, 0) node {$)$};
        \draw (7.1, 2) node {$\Bigg\Downarrow \textbf{Deletion}$};
      \end{scope}
    \end{tikzpicture}
  }
  \caption{Simplification of \nlang{``In 1964 Peter Higgs published his
      second paper in Physical Review Letters describing Higgs
      mechanism which predicted a new massive spin-zero boson for
      the first time.''}}\label{fig:example}
\end{figure*}

\subsection{Example Simplification}
\label{subsec:example}

Before describing the three main modules of our simplification
framework, we illustrate its working with an example. Figure
\ref{fig:example} shows the input semantic representation associated
with sentence (\ref{ex:derivation}C) and illustrates the successive
simplification steps yielding the intermediate and final simplified
sentences shown in (\ref{ex:derivation}S$_1$-S).

\begin{footnotesize}
\enumsentence{\label{ex:derivation} 
  {\bf C.} In 1964 Peter Higgs published his second paper in Physical
  Review Letters describing Higgs mechanism which predicted a new
  massive spin-zero boson for the first time.\\
  {\bf S$_1$ (Lex Simp).} In 1964 Peter Higgs wrote his second paper in
  Physical Review Letters explaining Higgs mechanism which predicted a
  new massive elementary particle for the first time.\\
  {\bf S$_2$ (Split).} In 1964 Peter Higgs wrote his second paper in Physical
  Review Letters explaining Higgs mechanism. Higgs mechanism predicted
  a new massive elementary particle for the first time.\\
  {\bf S (Deletion).} In 1964 Peter Higgs wrote his paper explaining
  Higgs mechanism.  Higgs mechanism predicted a new elementary
  particle.}
\end{footnotesize}

First, the input (\ref{ex:derivation}C) is rewritten as
(\ref{ex:derivation}S$_1$) by replacing standard
words with simpler ones using the context aware lexical simplification
method proposed in \cite{biran2011putting}. 

Splitting is then applied to the semantic representation of
(\ref{ex:derivation}S$_1$). Following \newcite{narayan2014hybrid}, we
use Boxer \footnote{\url{http://svn.ask.it.usyd.edu.au/trac/candc},
  Version 1.00} \cite{curran2007linguistically} to map the output
sentence from the lexical simplification step (here S$_1$) to a
Discourse Representation Structure (DRS, \cite{kamp81}). The DRS for
S$_1$ is shown at the top of Figure~\ref{fig:example} and a graph
representation\footnote{The DRS to graph conversion goes through
  several preprocessing steps: the relation \emph{nn} is inverted
  making modifier noun (\emph{higgs}) dependent of modified noun
  (\emph{mechanism}), \emph{named} and \emph{timex} are converted to
  unary predicates, e.g., $named(x,peter)$ is mapped to $peter(x)$ and
  $timex(x)=1964$ is mapped to $1964(x)$; and nodes are introduced for
  orphan words (e.g., \emph{which}).} of the dependencies between its
variables is shown immediately below.  In this graph, each DRS
variable labels a node in the graph and each edge is labelled with the
relation holding between the variables labelling its end vertices. The
two tables to the right of the picture show the predicates (top table)
associated with each variable and the relation label (bottom table)
associated with each edge. Boxer also outputs the associated positions
in the complex sentence for each predicate (not shown in the DRS but
shown in the graph tables). Orphan words i.e., words which have no
corresponding material in the DRS (e.g., \emph{which} at position
$16$), are added to the graph (node $O_1$) thus ensuring that the
position set associated with the graph exactly generates the input
sentence.

Using probabilities over sequences of thematic role sets acquired from
the DRS representations of SWKP, the split module determines where and
how to split the input DRS. In this case, one split is applied between
$X_{11}$ (\nlang{explain}) and $X_{10}$ (\nlang{predict}).  The
simpler sentences resulting from the split are then derived from the
DRS using the word order information associated with the predicates,
duplicating or pronominalising any shared element (e.g., \nlang{Higgs
  mechanism} in Figure~\ref{fig:example}) and deleting any Orphan
words (e.g., \nlang{which}) which occurs at the split
boundary. Splitting thus derives S$_2$ from S$_1$.

Finally, deletion or sentence compression applies transforming S$_2$
into S$_3$.

\subsection{Context-Aware Lexical Simplification}
\label{subsec:lexicalsimplification}

We extract context-aware lexical simplification rules from EWKP and
SWKP\footnote{We downloaded the snapshots of English Wikipedia dated
  2013-12-31 and of Simple English Wikipedia dated 2014-01-01
  available at \url{http://dumps.wikimedia.org}.} using the approach
described by \newcite{biran2011putting}. The underlying intuition
behind these rules is that the word $C$ from EWKP can be replaced with
a word $S$ from SWKP if $C$ and $S$ share similar contexts (ten token
window) in EWKP and SWKP respectively. Given an input sentence and the
set of simplification rules extracted from EWKP and SWKP, we then
consider all possible $(C,S)$ substitutions licensed by the extracted
rules and we identify the best combination of lexical simplifications
using dynamic programming and rule scores which capture the adequacy,
in context, of each possible substitution\footnote{For more details on
  the extraction of lexical simplification rules, we refer the reader
  to \newcite{biran2011putting}. For more details on the application
  of these rules using dynamic programming, we refer the reader to
  \newcite{narayan2014thesis}.}.

\subsection{Sentence Splitting}
\label{subsec:sentencesplitting}

A distinguishing feature of our approach is that splitting is
based on deep semantic representations rather than phrase structure
trees -- as in \cite{zhu2010monolingual,woodsend2011learning} -- or
dependency trees -- as in \cite{siddharthan-mandya:2014:EACL}.  

While \newcite{woodsend2011learning} report learning 438 splitting
rules for their simplification approach operating on phrase structure
trees \newcite{siddharthan-mandya:2014:EACL} defines 26 hand-crafted
rules for simplifying apposition and/or relative clauses in dependency
structures and 85 rules to handle subordination and coordination.  

In contrast, we do not need to specify or to learn complex rewrite
rules for splitting a complex sentence into several simpler sentences.
Instead, we simply learn the probability of sequences of thematic role
sets likely to cooccur in a simplified sentence.

The intuition underlying our approach is that:

\vspace{-0.2cm}
\begin{quote}{\footnotesize
Semantic representations give a clear handle on events, on their
associated roles sets and on shared elements thereby facilitating both
the identification of possible splitting points and the reconstruction
of shared elements in the sentences resulting from a split.}
\end{quote}
\vspace{-0.2cm}

For instance, the DRS in
Figure~\ref{fig:example} makes clear that sentence
(\ref{ex:derivation}S$_1$) contains 3 main events and that
\nlang{Higgs mechanism} is shared between two propositions.

To determine whether and where to split the input sentence, we use a
probabilistic model trained on the DRSs of the Simple Wikipedia
sentences and a language model also trained on Simple Wikipedia.
Given the event variables contained in the DRS of the input sentence,
we consider all possible splits between subsequences of events and
choose the split(s) with maximum split score. For instance, in the
sentence shown in Figure \ref{fig:example}, there are three event
variables $X_3$, $X_{10}$ and $X_{11}$ in the DRS. So we will consider
5 split possibilities namely, no split ($\{X_3, X_{10}, X_{11}\}$),
two splits resulting in three sentences describing an event each
($\{X_3\}$, $\{X_{10}\}$, $\{X_{11}\}$) and one split resulting in two
sentences describing one and two events respectively (i.e.,
$(\{X_3\},\{X_{10},X_{11}\})$, $(\{X_{3},X_{10}\},\{X_{11}\})$ and
$\{X_{10}\},\{X_3,X_{11}\}$).  The split $\{X_{10}\},\{X_3,X_{11}\}$
gets the maximum split score and is chosen to split the sentence
(\ref{ex:derivation}S$_1$) producing the sentences
(\ref{ex:derivation}S$_2$).

\begin{table}[htbp]
  \begin{center}
    \begin{footnotesize}
      \begin{tabular}{|l|r|}\hline
        Semantic Pattern & prob. \\ \hline
        \parbox{6cm}{$\langle$ ($agent$, $patient$) $\rangle$ } & $0.059$ \\
        \parbox{6cm}{$\langle$ ($agent$, $in$, $in$, $patient$)  $\rangle$} & $0.002$ \\
        \parbox{6cm}{$\langle$ ($agent$, $patient$), ($agent$, $in$, $in$,
          $patient$) $\rangle$} & $0.023$ \\
        \hline
      \end{tabular}
      \caption{\small Split Feature Table (SFT) showing some of the
        semantic patterns from Figure~\ref{fig:example}.
      }\label{tab:sft}
      \vspace{-0.3cm}
  \end{footnotesize}
  \end{center}
\end{table}

Formally, the split score $P_{split}$ associated with the splitting of
a sentence $S$ into a sequence of sentences $s_1 ... s_n$ is defined
as:

\begin{table*}[htbp]
  \begin{center}
    \begin{footnotesize}
      \begin{tabular}{|p{1.2cm}|rr|rr|r|r|r|r|}\hline
        
        \multirow{3}{*}{System} &
        \multicolumn{4}{|c|}{\parbox{4cm}{Levenshtein Edit distance}}
        & \multirow{3}{*}{\parbox{1.5cm}{BLEU \\ w.r.t simple}} &
        \multirow{3}{*}{\parbox{1.3cm}{Sentences \\ with splits}} &
        \multirow{3}{*}{\parbox{1.2cm}{Average sentence length}} &
        \multirow{3}{*}{\parbox{1.2cm}{Average token length}}
        \\ \cline{2-5}
        & \multicolumn{2}{|c|}{\parbox{2cm}{Complex to System}} &
        \multicolumn{2}{|c|}{\parbox{2cm}{System to Simple}} & & & &\\ \cline{2-5}
        & LD & No edit & LD & No edit & & &  & \\ \hline
        GOLD & 12.24 & 3 & 0 & 100 & 100 & 28 & 27.80 & 4.40 \\ 
        Zhu & 7.87 & 2 & 14.64 & 0 & 37.4 & 80 & 24.21 & 4.38\\
        Woodsend & 8.63 & 24 & 16.03 & 2 & 42 & 63 & 28.10 & 4.50\\
        Wubben & 3.33 & 6 & 13.57 & 2 & 41.4 & 1 & 28.25 & 4.41\\
        Narayan & 6.32 & 4 & 11.53 & 3 & 53.6 & 10 & 26.24 & 4.36\\\hline
        UNSUP & 6.75 & 3 & 14.29 & 0 & 38.47 & 49 & 26.22 & 4.40 \\\hline
      \end{tabular}
      \caption{\small Automatic evaluation results.  Zhu, Woodsend,
        Wubben, Narayan are the best output of the models of Zhu et
        al. (2010), Woodsend and Lapata (2011), Wubben et al. (2012)
        and Narayan and Gardent (2014) respectively. UNSUP is our
        model.}\label{tab:autoeval}
      \vspace{-0.5cm}
    \end{footnotesize}    
  \end{center}
\end{table*}

{
  \footnotesize
  \vspace{-0.4cm}
  \begin{equation*}\label{eq:split} P_{split} = \frac{1}{n} \sum_{s_i} 
    \frac{L_{split}}{L_{split}+\mid L_{split} - L_{s_i} \mid } \times
    lm_{s_i} \times SFT_{s_i}
  \end{equation*}
  \vspace{-0.3cm}
}

where $n$ is the number of sentences produced after splitting;
$L_{split}$ is the average length of the split sentences ($L_{split} =
\frac{L_S }{n}$ where $L_S$ is the length of the sentence $S$);
$L_{s_i}$ is the length of the sentence $s_i$; $lm_{s_i}$ is the
probability of $s_i$ given by the language model and $SFT_{s_i}$ is
the likelihood of the semantic pattern associated with $s_i$. The
Split Feature Table (SFT, Table~\ref{tab:sft}) is derived from the
corpus of DRSs associated with the SWKP sentences and the counts of
sequences of thematic role sets licenced by the DRSs of SWKP
sentences. Intuitively, $P_{split}$ favors splits involving frequent
semantic patterns (frequent sequences of thematic role sets) and
sub-sentences of roughly equal length. This way of semantic pattern
based splitting also avoids over-splitting of a complex sentence.



\subsection{Phrasal Deletion}
\label{subsec:deletion}

Following \newcite{filippova2008dependency}, we formulate phrase
deletion as an optimization problem which is solved using integer
linear programming\footnote{In our implementation, we use $lp\_solve$,
  \url{http://sourceforge.net/projects/lpsolve}. }. Given the DRS $K$
associated with a sentence to be simplified, for each relation $r\in
K$, the deletion module determines whether $r$ and its associated DRS
subgraphs should be deleted by maximising the following objective
function:

{
  \footnotesize
  \vspace{-0.15cm}
  \begin{equation*}
    \sum_{x} x^r_{h,w} \times P(r|h) \times P(w) \quad r \not\in
    \{agent, patient, theme, eq\}
  \end{equation*}
  \vspace{-0.25cm}
}

where for each relation $r\in K$, $x^r_{h,w} = 1$ if $r$ is preserved
and $x^{r}_{h,w} = 0$ otherwise; $P(r|h)$ is the conditional
probability (estimated on the DRS corpus derived from SWKP) of $r$
given the head label $h$; and $P(w)$ is the relative frequency of $w$
in SWKP\footnote{To account for modifiers which are represented as
  predicates on nodes rather than relations, we preprocess the DRSs
  and transform each of these predicates into a single node subtree of
  the node it modifies. For example in Figure \ref{fig:example},
  the node $X_2$ labeled with the modifier predicate \nlang{second} is
  updated to a new node $X'_2$ dominating a child labeled with that
  predicate and related to $X'_2$ by a modifier relation.}.

Intuitively, this objective function will favor obligatory
dependencies over optional ones and simple words (i.e., words that are
frequent in SWKP). In addition, the objective function is subjected to
constraints which ensure (i) that some deletion takes place and (ii)
that the resulting DRS is a well-formed graph.





\begin{table*}[htbp]
  \begin{center}
    \begin{footnotesize}
      \begin{tabular}{|p{4.3cm}|rr|rr|rr|r|r|}\hline

        \multirow{3}{*}{System} &
        \multicolumn{4}{|c|}{\parbox{4cm}{Levenshtein Edit distance}}
        & \multicolumn{2}{|c|}{\multirow{2}{*}{\parbox{2cm}{BLEU
              Scores \\ with respect to}}} &
        \multirow{3}{*}{\parbox{1.2cm}{Average sentence length}} &
        \multirow{3}{*}{\parbox{1.2cm}{Average token
            length}}\\ \cline{2-5}

        & \multicolumn{2}{|c|}{\parbox{2cm}{Complex to System}} &
        \multicolumn{2}{|c|}{\parbox{2cm}{System to Simple}} & & & &
        \\ \cline{2-7}
        
        & LD & No edit & LD & No edit & complex & simple & &  \\ \hline
        
        complex & 0 & 100 & 12.24 & 3 & 100 & 49.85 & 27.80 & 4.62 \\
        LexSimpl & 2.07 & 22  & 13.00 & 1 & 82.05 & 44.29 & 27.80 & 4.46 \\
        Split & 2.27 & 51 & 13.62 & 1 & 89.70 & 46.15 & 29.10 & 4.63 \\
        Deletion & 2.39 & 4 & 12.34 & 0 & 85.15 & 47.33 & 25.41 & 4.54 \\
        LexSimpl-Split & 4.43 & 11 & 14.39 & 0 & 73.20 & 41.18 & 29.15 & 4.48 \\
        LexSimpl-Deletion & 4.29 & 3 & 13.09 & 0 & 69.84 & 41.91 & 25.42 & 4.38 \\
        Split-Deletion & 4.63 & 4 & 13.42 & 0 & 77.82 & 43.44 & 26.19 & 4.55 \\
        LexSimpl-Split-Deletion & 6.75 & 3 & 14.29 & 0 & 63.41 & 38.47 & 26.22 & 4.40 \\
        GOLD (simple) & 12.24 & 3 & 0 & 100 & 49.85 & 100 & 23.38 & 4.40 \\ \hline 

      \end{tabular}
      \caption{\small Automated Metrics for Simplification: Modular
        evaluation. LexSimpl-Split-Deletion is our final system
        UNSUP.}\label{tab:modular}
      \vspace{-0.5cm}
    \end{footnotesize}
  \end{center}
\end{table*}

\section{Evaluation}
\label{sec:xps}

We evaluate our approach both globally and by module focusing in
particular on the splitting component of our simplification approach.

\subsection{Global evaluation}

The testset provided by \newcite{zhu2010monolingual} was used by four
supervised systems for automatic evaluation using metrics such as
BLEU, sentence length and number of edits. In addition, most recent
simplification approaches carry out a human evaluation on a small set
of randomly selected complex/simple sentence pairs. Thus
\newcite{wubben2012sentence}, \newcite{narayan2014hybrid} and
\newcite{siddharthan-mandya:2014:EACL} carry out a human evaluation on
20, 20 and 25 sentences respectively.

Accordingly, we perform an automatic comparative evaluation using
\cite{zhu2010monolingual}'s testset namely, an aligned corpus of
$100/131$ EWKP/SWKP sentences; and we carry out a human-based
evaluation.




\paragraph{Automatic Evaluation}

Following \newcite{wubben2012sentence}, \newcite{zhu2010monolingual}
and \newcite{woodsend2011learning}, we use metrics that are directly
related to the simplification task namely, the number of splits in the
overall data, the number of output sentences with no edits (i.e.,
sentences which have not been simplified) and the average Levenshtein
distance (LD) between the system output and both the complex and the
simple reference sentences. We use BLEU\footnote{Moses support tools:
  multi-bleu \url{http://www.statmt.org/moses/?n=Moses.SupportTools}.}
as a means to evaluate how close the systems output are to the
reference corpus.

Table~\ref{tab:autoeval} shows the results of the automatic
evaluation.  The most noticeable result is that our unsupervised
system yields results that are similar to those of the supervised
approaches.

The results also show that, in contrast to Woodsend system which often
leaves the input unsimplified (24\% of the input), our system almost
always modifies the input sentence (only 3\% of the input are not
simplified); and that the number of simplifications including a split
is relatively high (49\% of the cases) suggesting a good ability to
split complex sentences into simpler ones.

\paragraph{Human Evaluation}
Human judges were asked to rate input/output pairs w.r.t. to adequacy
(How much does the simplified sentence(s) preserve the meaning of the
input?), to simplification (How much does the generated sentence(s)
simplify the complex input?) and to fluency (how grammatical and
fluent are the sentences?).

We randomly selected 18 complex sentences from Zhu's test corpus and
included in the evaluation corpus: the corresponding simple (Gold)
sentence from Zhu's test corpus, the output of our system (UNSUP) and
the output of the other four systems (Zhu, Woodsend, Narayan and
Wubben) which were provided to us by the system authors\footnote{We
  upload the outputs from all the systems as supplementary material
  with this paper.}.  We collected ratings from 18 participants. All
were either native speakers or proficient in English, having taken
part in a Master taught in English or lived in an English speaking
country for an extended period of time.  The evaluation was done
online using the LG-Eval toolkit
\cite{kow2012lg}\footnote{http://www.nltg.brighton.ac.uk/research/lg-eval/}
and a Latin Square Experimental Design (LSED) was used to ensure a
fair distribution of the systems and the data across raters.

\begin{table}[htbp]
  \begin{center}
    \begin{footnotesize}
      \vspace{-0.2cm}
      \begin{tabular}{|l|r|r|r|}\hline
        Systems & Simplicity & Fluency & Adequacy \\ \hline
        GOLD & 3.62 & 4.69 & 3.80\\
        Zhu & 2.62 & 2.56 &  2.47 \\
        Woodsend & 1.69 & 3.15 & 3.15 \\
        Wubben & 1.52 & 3.05 & 3.38 \\
        Narayan & 2.30 & 3.03 & 3.35 \\ \hline
        UNSUP & 2.83 & 3.56 & 2.83\\ \hline
      \end{tabular}
      \caption{\small Average Human Ratings for simplicity, fluency
        and adequacy.}\label{tab:humaneval}
      \vspace{-0.3cm}
    \end{footnotesize}
  \end{center}
\end{table}

\begin{table*}[htbp]
\begin{center}
\begin{footnotesize}
\begin{tabular}{|lr|r|r|r|r|r|r|}\hline

\multicolumn{2}{|c|}{System pairs} & \multicolumn{6}{|c|}{Average
  Score (number of split sentences)} \\ \cline{3-8}

\multirow{2}{*}{A} & \multirow{2}{*}{B} &
\multirow{2}{*}{ALL-A} & \multirow{2}{*}{ALL-B} &
\multirow{2}{*}{ONLY-A} & \multicolumn{2}{c|}{BOTH-AB} &
\multirow{2}{*}{ONLY-B} \\ \cline{6-7}

& & & & & A & B & \\ \hline

\multirow{5}{*}{UNSUP} & GOLD & \multirow{5}{*}{2.37(49)} & 3.85(28) &
2.15(32) & 2.80(17) & 3.70(17) & 4.05(11) \\

& Zhu & & 2.25(80) & 1.53(4) & 2.45(45) & 2.42(45) & 2.02(35)\\
& Woodsend & & 2.08(63) & 2.42(11) & 2.36(38) & 2.29(38) & 1.78(25) \\
& Wubben & & 2.73(1) & 2.32(48) & 4.75(1) & 2.73(1) & 0(0) \\
& Narayan & & 2.09(10) & 2.29(41) & 2.78(8) & 1.79(8) & 3.81(2) \\\hline
\end{tabular}
\caption{\small Pairwise split evaluation: Each row shows the pairwise
  comparison of the quality of splits in UNSUP and some other
  system. Last six columns show the average scores and number of
  associated split sentences. The second column (ALL-A) and the third
  column (ALL-B) present the quality of all splits by systems A and B
  respectively. The fourth column (ONLY-A) represents sentences where
  A splits but not B. The fifth and sixth columns represents sentences
  where both A and B split. The seventh column (ONLY-B) represents
  sentences where B splits but not
  A.}\label{pairwise-split-evaluation}
\vspace{-0.5cm}
\end{footnotesize}
\end{center}
\end{table*}




Table~\ref{tab:humaneval} shows the average ratings of the human
evaluation on a scale from 0 to 5. Pairwise comparisons between all
models and their statistical significance were carried out using a
one-way ANOVA with post-hoc Tukey HSD tests. If we group together
systems for which there is no significant difference (significance
level: p $<$ 0.05), our system is in the first group together with
Narayan and Zhu for simplicity; in the first group for fluency; and in
the second group for adequacy (together with Woodsend and Zhu).  A
manual examination of the results indicates that UNSUP achieves good
simplicity rates through both deletion and sentence splitting. Indeed,
the average word length of simplified sentences is smaller for UNSUP
(26.22) than for Wubben (28.25) and Woodsend (28.10); comparable with
Narayan (26.19) and higher only than Zhu (24.21).

\vspace{-0.05cm}
\subsection{Modular Evaluation}

To assess the relative impact of each module (lexical simplification,
deletion and sentence splitting), we also conduct an automated
evaluation on each module separately. The results are shown in
Table~\ref{tab:modular}.

One first observation is that each module has an impact on
simplification. Thus the average Levenshtein Edit distance (LD) to
the source clause (complex) is never null for any module while the
number of ``No edit'' indicates that lexical simplification modifies
the input sentence in 78\%, sentence splitting 49\% and deletion 96\%
of the cases.  

In terms of output quality and in particular, similarity with respect
to the target clause, deletion is the most effective (smallest LD,
best BLEU score w.r.t. target). Further, the results for average
token length indicate that lexical simplification is effective in
producing shorter words (smaller average length for this
module compared to the other two modules).

Predictably, combining modules yields systems that have stronger
impact on the source clause (higher LD to complex, lower number of No
Edits) with the full system (i.e., the system combining the 3
modules) showing the largest LD to the sources (LD to complex) and the
smallest number of source sentences without simplification (3 No
Edits). 

\vspace{-0.1cm}
\subsection{Sentence Splitting Using Deep Semantics}

\begin{table*}[htbp]
  \begin{footnotesize}
    \begin{tabular}{|l|p{15cm}|}\hline
      S1 &  {\bf Complex.} This array distributes data across multiple disks,
      but the array is seen by the computer user and operating system as
      one single disk. \\
      &  {\bf Zhu.} This array sells data across multiple disks but the array
      is seen. The computer user and operating as one disk. [0]\\
      &  {\bf Woodsend.} This array distributes data across several disks.
      The array is seen by the computer user and operating system as one
      single disk. [4.85]\\
      &  {\bf UNSUP.} This array moves data across disks. The array is seen
      by the computer user and operating system as one disk. [4.83]\\\hline
      S2 &   {\bf Complex.} Here, simulations of evolution using evolutionary
      algorithms and artificial life started with the work of Nils Aall
      Barricelli in the 1960s, and was extended by Alex Fraser, who
      published a series of papers on simulation of artificial
      selection. \\
      & {\bf Zhu.} Here, simulations of evolution using evolutionary
      algorithms and artificial life started with the work of nils aall
      barricelli in the 1960s, and was extended by alex fraser. Alex
      fraser published a series of papers on feeling of artificial
      selection. [3.6]\\
      & {\bf Woodsend.} It has Genetic algorithms. Evolution is using
      artificial life. This is Nils Aall Barricelli in the 1960s. Here,
      simulations started with the work. He was extended by Alex
      Fraser. He published a group of papers on simulation of artificial
      selection. [0.33]\\
      & {\bf UNSUP.} Here, models of evolution using evolutionary algorithms
      and artificial life started with the work in the 1960s, and was led
      by Alex Fraser.  Alex Fraser wrote a series of papers on model of
      selection. [5]
      \\\hline
      S3 &  {\bf Complex.} By 1928, the regional government was moved from the old
      Cossack capital Novocherkassk to Rostov, which also engulfed the
      nearby Armenian town of Nor Nakhijevan. \\
      &  {\bf Zhu.} By 1928, the government was moved from the old cossack
      capital novocherkassk to rostov. Rostov also of the city the nearby
      armenian town of nor nakhijevan. [2.8]\\
      & {\bf Woodsend.} By 1928, the regional government was moved from the
      old Cossack capital Novocherkassk to Rostov. Both also engulfed the
      nearby Armenian town of Nor Nakhijevan. [3]\\
      & {\bf Wubben.} by 1928 , the regional government was moved from the
      old cossack capital novocherkassk to rostov. the nearby armenian
      town of nor nakhichevan. [2.7]\\
      & {\bf Narayan.} by 1928, the regional government was moved from the
      old cossack capital novocherkassk to rostov. rostov that engulfed
      the nearby armenian town of nor nakhichevan. [2.7]\\
      &  {\bf UNSUP.} The regional government was moved from the old Cossack
      capital Novocherkassk to Rostov. Rostov also absorbed the nearby
      town of Nor Nakhijevan. [4.75]
      \\\hline
    \end{tabular}
  \end{footnotesize}
  \caption{\small Example Outputs for  Sentence splitting with their
    average human annotation scores.}\label{fig:splits}
  \vspace{-0.4cm}
\end{table*}

To compare our sentence splitting approach with existing systems, we
collected in a second human evaluation, all the outputs for which at
least one system applied sentence splitting. The raters were then
asked to compare pairs of split sentences produced by two distinct
systems and to evaluate the quality (0:very bad to 5:very good) of
these split sentences taking into account boundary choice, sentence
completion and sentence reordering. 

Table~\ref{pairwise-split-evaluation} shows the results of this second
evaluation. For each system pair comparing UNSUP (A) with another system (B),
the Table gives the scores and the number of splits of both systems:
for the inputs on which both systems split (BOTH-AB), on which only
UNSUP splits (ONLY-A) and on which only the compared system split
(ONLY-B).

UNSUP achieves a better average score (ALL-A = 2.37) than all other
systems (ALL-B column) except Wubben (2.73). However Wubben only
achieves one split and on that sentence, UNSUP score is 4.75 while
Wubben has a score of 2.73 and produces an incorrect split (cf. $S_3$
in Figure~\ref{fig:splits}). 
UNSUP

In terms of numbers of splits, three systems often simplify by
splitting namely Zhu (80 splits), Woodsend (63) and UNSUP (49).
Interestingly, Narayan, trained on the parallel corpus of Wikipedia
and Simplified Wikipedia splits less often (10 splits vs 49 for UNSUP)
and less well (2.09 average score versus 2.37 for UNSUP). This is
unsurprising as the proportion of splits in SWKP was reported in
\cite{narayan2014hybrid} to be a low 6\%. In contrast, the set of
observations we use to learn the splitting probability is the set of
all sequences of thematic role sets derived from the DRSs of the SWKP
corpus.

In sum, the unsupervised, semantic-based splitting strategy allows for
a high number (49\%) of good quality (2.37 score) sentence splits .
Because there are less possible patterns of thematic role sets in
simple sentences than possible configurations of parse/dependency
trees for complex sentences, it is less prone to data sparsity than
the syntax based approach. Because the probabilities learned are not
tied to specific syntactic structures but to more abstract semantic
patterns, it is also perhaps less sensitive to parse errors.

\subsection{Examples from the Test Set}

Table~\ref{fig:splits} shows some examples from the evaluation
dataset which were selected to illustrate the workings of our approach
and to help interpret the results in
Table~\ref{tab:autoeval}, \ref{tab:humaneval}
and~\ref{pairwise-split-evaluation}.

S1 and S2 and S3 show examples of context-aware unsupervised lexical
substitutions which are nicely performed by our system. In S1,
\nlang{The array \underline{distributes} data} is correctly simplified
to \nlang{The array \underline{moves} data} whereas
Zhu's system incorrectly simplifies this clause to \nlang{The array
  \underline{sells} data}. Similarly, in S2, our system correctly simplifies
\nlang{Papers on \underline{simulation} of artificial selection} to
\nlang{Papers on \underline{models} of selection} while the other
systems either do not simplify or simplify to \nlang{Papers on
  \underline{feeling}}.


For splitting, the examples show two types of splitting performed by
our approach namely, splitting of coordinated sentences (S1) and
splitting between a main and a relative clause (S2,S3). S2 illustrates
how the Woodsend system over-splits, an issue already noticed in
\cite{siddharthan-mandya:2014:EACL}; and how Zhu's system predicts an
incorrect split between a verb (\nlang{seen}) and its agent argument
(\nlang{by the user}). Barring a parse error, such incorrect splits
will not be predicted by our approach since, in our cases, splits only
occur between (verbalisations of) events. S1, S2 and S3 also
illustrates how our semantic based approach allows for an adequate
reconstruction of shared elements.

\vspace{-0.1cm}
\section{Conclusion}
\label{sec:conclusion}

A major limitation for supervised simplification systems is the
limited amount of available parallel standard/simplified data.  In
this paper, we have shown that it is possible to take an unsupervised
approach to sentence simplification which requires a large corpus of
standard and simplified language but no alignment between the
two. This allowed for the implementation of contextually aware
substitution module; and for a simple, linguistically principled
account of sentence splitting and shared element reconstruction. 

\section{Acknowledgements}

We are grateful to Zhemin Zhu, Kristian Woodsend and Sander Wubben for
sharing their data.  We would like to thank our annotators for
participating in our human evaluation experiments and to anonymous
reviewers for their insightful comments. This research was supported
by an EPSRC grant (EP/L02411X/1) and an EU H2020 grant
(688139/H2020-ICT-2015; SUMMA). We also thank the French National
Research Agency for funding the research presented in this paper in
the context of the WebNLG project.



\bibliographystyle{acl}

\bibliography{unsupervised-simplification-reference}

\end{document}